\documentclass[runningheads]{llncs}
\usepackage[T1]{fontenc}
\usepackage{graphicx}
\usepackage{booktabs}
\usepackage{multirow}
\usepackage{amsmath}
\usepackage{microtype}
\usepackage{tcolorbox}
\usepackage{xcolor}

\begin{document}
\title{Ontology-Driven Control of Conversational Agents with a Use-Case for Job Interview Practice}
\titlerunning{Ontology-Driven Control of Conversational Agents}
\author{Barbara Gendron\inst{1} \and
Gaël Guibon\inst{1,2} \and
Mathieu d'Aquin\inst{1}}
\authorrunning{Gendron, Guibon and d'Aquin}

\institute{Université de Lorraine, CNRS, LORIA, Nancy France \and
Université Sorbonne Paris Nord, LIPN, CNRS, UMR 7030, F-93430, Villetaneuse, France\\
\email{{barbara.gendron, gael.guibon, mathieu.daquin}@loria.fr}}
\maketitle              
\begin{abstract} 

Large Language Models (LLMs) have been widely used as conversational agents in both domain-specific and open-domain situations. In domain-specific context, it is often necessary to adapt those LLMs to obtain conversations with desired characteristics, whether those apply at the level of each response (e.g. for tone or style) or more globally (i.e. at the level of a conversational strategy). Such adaptations are, however, generally realized in an ad hoc way, through specific, non-reusable, and non-transparent developments. In this paper, we propose an ontology-based approach to design conversational strategies that rely on the desired and measured characteristics of textual utterances in the conversation. Through a use case related to job interview practice, we show how this approach can be used to control LLMs to follow the strategy defined in the ontology. To this end, we automatically fine-tune the LLM to improve its compliance with ontologically-defined conversation characteristics. Our results show that training the LLM from formally defined concepts effectively enhances alignment with the requested characteristics. Hence, this work demonstrates how ontology-guided conversation control can be achieved while taking into account multiple, concurrent conversation aspects that are formally and transparently defined.

\end{abstract}
\section{Introduction}

Nowadays, conversational agents (CA) are in everyday life leading to question the impact of their constant usage~\cite{hennekeuser2024don}. The need for knowledge-enhanced CA arises as agents need knowledge to adapt to different use cases and contexts of many types~\cite{erickson2025llm}. It is now essential to be able to manipulate and control this knowledge especially for domain-specific conversations~\cite{city34054}. Such domains are customer support for which recent work proposed a model-friendly knowledge representation~\cite{su-etal-2025-llm}, CA dedicated to the medical domain with appropriate knowledge and behavior~\cite{cho-etal-2023-integrative}, or even in human resources for which a recent dataset was proposed to cope with this limitation~\cite{xu-etal-2024-hr}, to name a few. Even open-domain conversations require control on the content since they are more versatile. The recent first workshop on simulating conversational intelligence in chat~\cite{graham-etal-2024-findings} is evidence of this need.

In this work, we want to put external control on any domain conversational agents in order to make them follow a specified flow and adhere to given, well-defined constraints. To achieve this, it is important to be able to leverage general pretrained Large Language Models (LLMs)~\cite{chiang-etal-2022-recent} without requiring new domain-specific data to fine-tune them. Indeed, although LLMs are black boxes, in many use cases it is desirable to have predictable outputs, or at least predictable behavior, throughout the conversation. To be transparent, such predictable behavior should be defined formally, in a way external to the LLM itself. This definition, which we call a conversation strategy, might refer to multiple specific aspects of conversations, from topic to style and tone, and the agent might need to learn to respond to some of those aspects, especially when they were absent or considered differently in the LLM's original training. 
Our main hypothesis is that we can logically define such strategies, corresponding aspects, and the constraints in order to align and constrain the agent's behavior. Thus, we aim to control the flow of the conversation as well as guiding the training process of the LLM.

We address two main Research Questions to validate this hypothesis. First, (RQ1) we aim to verify whether and how conversation strategies can be ontologically defined in a way that takes multiple, concurrent aspects into account. They have to be usable to derive at every turn the control instructions to be sent to the LLM. Those strategies should naturally, at least partially, be based on measuring some aspects of the current conversation and adapt to those aspects. This means also that both the content and other components (style, tone, etc.) of the conversation should be controllable ontology-based instructions. 
Second, (RQ2) we verify whether the compliance of an LLM to ontology-defined aspects of the conversation can be improved through fine-tuning. Such fine-tuning focuses on linguistic and affective elements of the conversation that are not straightforwardly controllable thanks to the pre-training of the LLM alone. It should not require dedicated domain-specific data but rely on the ontology to automatically annotate a generated dataset during training. Therefore, our contribution is threefold: 
\begin{enumerate}
    \item To the best of our knowledge, we propose the first work on conversational strategies defined in an ontology.
    \item We show how an ontology can serve to control open-domain conversational agents.
    \item We show first results in improving the LLM's compliance with instruction from the ontology through fine-tuning. 
\end{enumerate}

As no comparable approaches, tasks, or datasets were found in the literature, we designed a suitably complex use case to illustrate and validate our method: A conversational agent that is adapted, thanks to ontology control, to support users in practicing for job interviews. We publicly share our code, data, and ontology\footnote{\url{https://github.com/B-Gendron/OntoDrivenConversation}}.

\section{Related Work}

We divide the related work into three different areas: conversation strategies, controlled generation, and the combination of LLMs and ontologies. While they all contribute to the foundation of our work, they have not had much intersection before now.

\subsection{Conversation Strategies}

Conversation strategies start in human interaction and how to automatically recognize them, especially focusing on strategies that help socially linking humans by building, maintaining, or destroying relationships. Researchers attempted to code the strategies using different types of conversation strategies, namely: ``\textit{Self-Disclosure}'', ``\textit{Referring to a shared experience}'', ``\textit{Praise}'', and ``\textit{Violation of Social Norms}''~\cite{zhao-etal-2016-automatic}. Those are still considered valid types of conversation strategies for chatbots in recent work~\cite{liang2024dialoging}. 
Conversation strategies can also be considered predictors, such as for helping resisting strategies in persuasive conversations~\cite{dutt-etal-2021-resper}. While this can be linked to our job interview use-case, strategy-enhanced role-playing using LLMs represents a novel emerging approach to exploiting conversational strategies in many scenarios. SweetieChat~\cite{ye-etal-2025-sweetiechat} is one example and identifies three roles in their conversation strategy. However, they constructed a specific dataset for the fine-tuning of the language model. The recent Strategy-Relevant Attention metric also requires new data on emotional support conversations~\cite{madani-srihari-2025-steering}. In our work, we do not require a specific dataset: the ontology we design and exploit prevents this need.

Proactive conversational agents, whether they are task-oriented, target-guided, or negotiation-focused, to name a few, are types of conversational agents leveraging conversational strategies. These proactive conversational agents mainly consider the strategy through a prompt integration~\cite{deng2024towards}, thus doing In-Context Learning~\cite{brown2020language}. However, recent research considers LLM inference for generating and reusing strategies. Mainly linked to planning~\cite{valmeekam2023planning}, conversational strategies are used for task adaptation with a persistent and dynamically generated textual memory~\cite{majumder2023clin} or by using LLM to hypothesize a world description~\cite{pmlr-v202-nottingham23a}. Somewhat close to role-playing, motivational interviewing is another research topic that requires strategies to align LLMs using In-Context Learning~\cite{brown2020language} to the dialogue behavior of Motivational Interviewers. They use LLMs to generate a set of natural language dialogue strategy descriptions, then use them as instructions at inference time~\cite{xie-etal-2024-shot-dialogue}. Contrary to our work, the inference is done using the LLMs with demonstration samples, while in our case we use an ontology to design the strategy.

In all these works, evaluation of conversation strategies is considered ad-hoc, through the evaluation of the final task, thus, ignoring the three proposed metrics for evaluating the conversation quality~\cite{yu2016strategy}: turn-level appropriateness, conversational depth, and information. This is due to the first two requiring human annotation while the last one is only based on new unique words, losing importance considering LLMs now. This is why in our work we do not focus on evaluating the conversation strategies, but mainly the control of the LLM through the ontology-defined strategy. Moreover, all these works require an interactive environment while we fully rely on an ontology and its individuals/instances.

\subsection{Controlled Generation and Behavior}

Since the end of the 2010s, much research has been conducted on the controllability of language models, especially large language models. Early works focused on training LLMs using control codes~\cite{hu2017toward} with Variational Auto-Encoders~\cite{kingma2013auto}. Later, researchers focused on conditional pretraining of a language model, for instance using over 50 control codes with conditional probability from the initial training phase~\cite{keskar:2019}. While this represents the beginning of conditional training for control, the control of the generation has also been proposed through additional smaller discriminators used to guide the fine-tuning of an already pretrained LLM, namely Plug-and-Play Language Model (PPLM)~\cite{dathathri2019plug}. This inspired research to extend PPLM with dedicated control codes to create a language model with controllable affective state~\cite{goswamy2020adapting}. In our work, the polarity and emotional load are incorporated into the ontology, thus we do not rely solely on control codes.

Recently, research works switched from only relying on control codes to either consider a distributional approach to control the generation~\cite{khalifa2021distribcontrol} or actually making LLMs follow instructions~\cite{ouyang2022training}, leading to the surge of instruction tuning studies such as for instruction controllable summarization~\cite{liu2024benchmarking}. This created the need for benchmarks of many types. Thus, appeared benchmarks and studies for controllability focusing on evaluating the effect of the instruction through a provided set of verifiable instructions~\cite{zhou2023instruction}. Also, Sun \textit{et al.}~\cite{sun-etal-2023-evaluating} proposed a benchmark on numerical constraints related to planning control and tried ten other benchmarks on story, rationale, paraphrase and constrained content generation. They conclude that LLMs struggle following fine-grained hard constraints, and identified paraphrase generation as very challenging. This is important with regards to our work since our use-case of job interview practice is a mix of constrained content generation, story generation (and role-playing), as well as paraphrase (very important in interviews). Additionally, the COLLIE benchmark tackled constrained text generation through a grammar-based framework~\cite{yao2024collie}, while the CELLO benchmark verifies LLM controllability over complex instructions~\cite{he2024can}, considering eight features for complex instructions: multi-tasking, semantics, format, quantity, heterogeneous information, long context, noise, and multi-turn. Finally, another study focuses on controlled text generation with LLMs for Japanese~\cite{kurihara2025lctg} and tried to propose a unified framework to evaluate their unified approach versus task-specific metrics. However, they only consider three tasks that do not fit our use-case, namely: Summarization, Ad Text Generation, Pros and Cons Generation.

\begin{figure*}[t]
    \centering
    \includegraphics[width=0.98\linewidth]{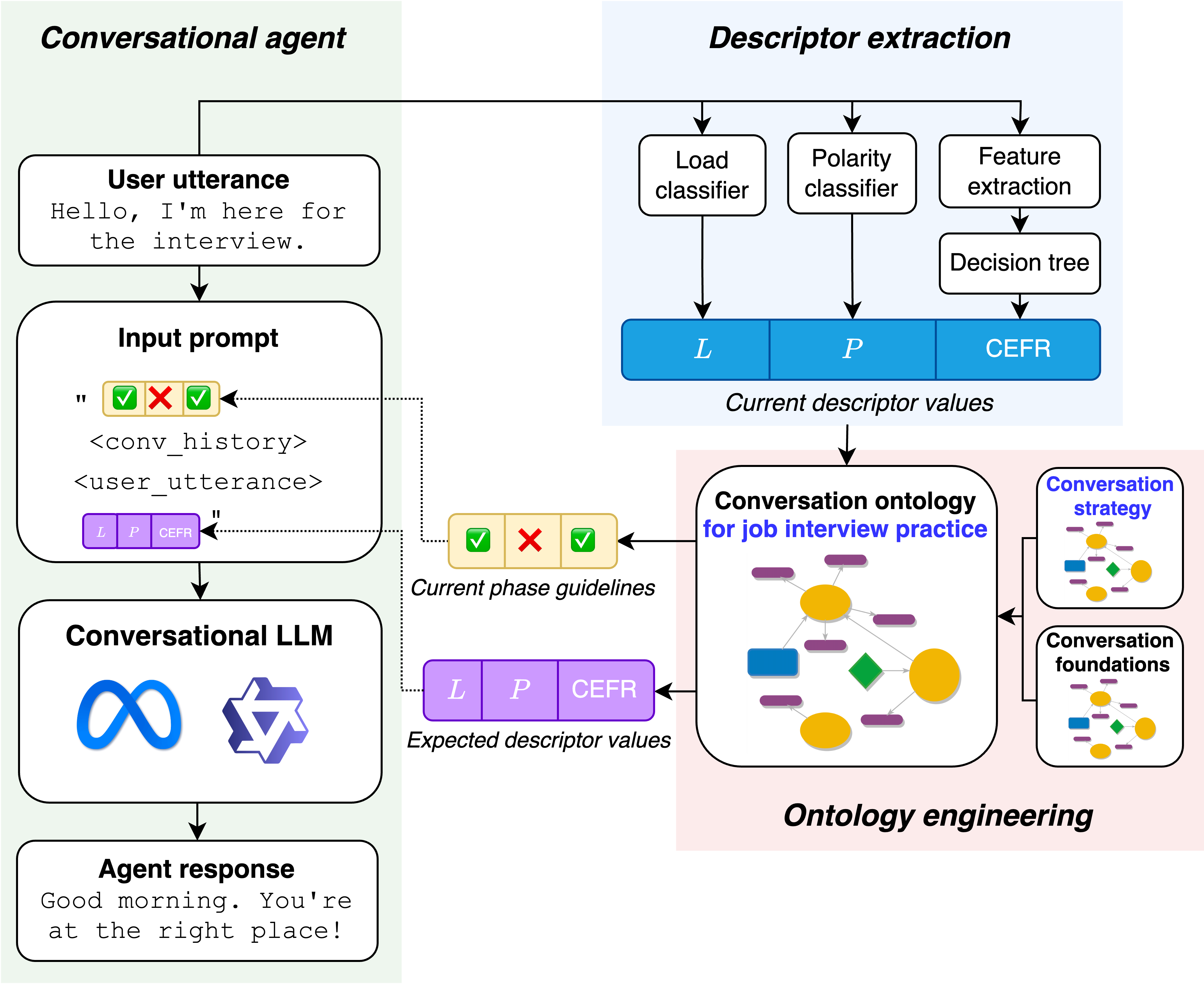}
    \caption{\centering A detailed overview of our system: we propose a knowledge-enhanced conversational agent, built on an LLM, that interacts with our own ontology to receive generation guidelines. These guidelines are conditioned on descriptor values extracted from the user utterance.}
    \label{fig:overall}
\end{figure*}

To sum up these control approaches, we can distinguish two types of controls, following Pascual \textit{et al.}~\cite{pascual2021plug}: the soft control (i.e., the general topic or mood) and the hard control (i.e., explicit constraints). In our use-case, we evaluate both a soft control of the LLM with the general orientation of the dialog and a hard control with job-specific elements and information. Moreover, as far as we know, no benchmark or study is considering a formal definition of the targeted task, which we provide with our ontology. To verify our approach, we gather the control elements within the ontology; thus, we only rely on control codes that stem from our ontology's reasoner. Finally, by contrast to these studies that attempt to automatically assess controllability, we consider a human evaluation to assist in evaluating our approach, enabling both soft and hard control evaluation.

\subsection{Knowledge-Enhanced LLM Conversational Agents}

While there has been a significant amount of work regarding hybrid approaches combining LLMs and knowledge graphs~\cite{agrawal-etal-2024-knowledge,perozzi2024letgraphtalkingencoding,Pan_2024} or ontologies~\cite{he2023exploring,giglou:2023,ekawclinical}, we focus on the case of knowledge-enhanced conversational agents. Such a hybridization setup is suitable for dialogue systems since conversational tasks often require complex knowledge management~\cite{kang2023knowledgegraphaugmentedlanguagemodels}. In that sense, there has been work towards the direction of enriching domain-specific conversational agents with factual, expert knowledge. It ranges from graph Question Answering (QA)~\cite{fatemi:2023} and Knowledge-Grounded QA~\cite{sun-etal-2022-jointlk,Wu2023RetrieveRewriteAnswerAK} to Domain Adaptation tasks~\cite{sreedhar-parisien-2022-prompt}. 

Furthermore, in the direction of developing conversational strategies, hybridization contributes significantly to planning tasks, as demonstrated by Gloria Silva~\textsl{et al.}~\cite{gloria-silva-etal-2024-plan}. Then, besides domain-specific conversations, knowledge enhancement can also be suitable for open-domain conversations, therefore focusing on more foundational aspects such as response relevance, language level, or emotions. Specifically, Varshney~\textsl{et al.}~\cite{varshney:2023} emphasizes that understanding human emotions is crucial for conversational models, arguing that emotionally aware agents improve user experience by facilitating meaningful dialogues that recognize and respond to emotions. In that sense, Kang \textsl{et al.}~\cite{kang2023knowledgegraphaugmentedlanguagemodels} proposes a knowledge-enhanced dialogue system that appeared more relevant to human evaluators due to its context-awareness enabled by subgraph retrieval. However, such approaches using the interaction between the LLM and external knowledge to implement conversational control often result in complex system architectures, such as that proposed by Varshney \textsl{et al}~\cite{Varshney2023-emokbgan}. 

In our case, we propose a hybrid LLM/ontology framework where LLM outputs are guided by an ontologically defined conversation strategy. In addition to ensuring a coherent and predictable conversation flow thanks to its formally defined strategy, our approach is lightweight and conceptually simpler, which allows for both greater efficiency and improved interoperability in conversational control. More precisely, at each turn of speech, the ontology can be used to reason on both the current utterance and the global state of the conversation, in order to deduce the instructions to be given to the conversational agent at that point in the dialogue.

\section{Approach}
\label{sec:approach}

Figure~\ref{fig:overall} provides a detailed description of the system with two main components: the first is the conversational agent, based on an LLM, that operates the discussion with the user. Then, to enable control on this conversational agent, we propose an ontology-based pipeline resulting in a conversation ontology tailored for our specific use-case, i.e. job interview practice. In order for the conversation ontology to provide guidelines as the outcome of the reasoning process, we need to specify the descriptor values that characterize the current utterance. This motivates the descriptor extraction part.

\begin{figure*}[ht!]
    \centering
    \includegraphics[width=\linewidth]{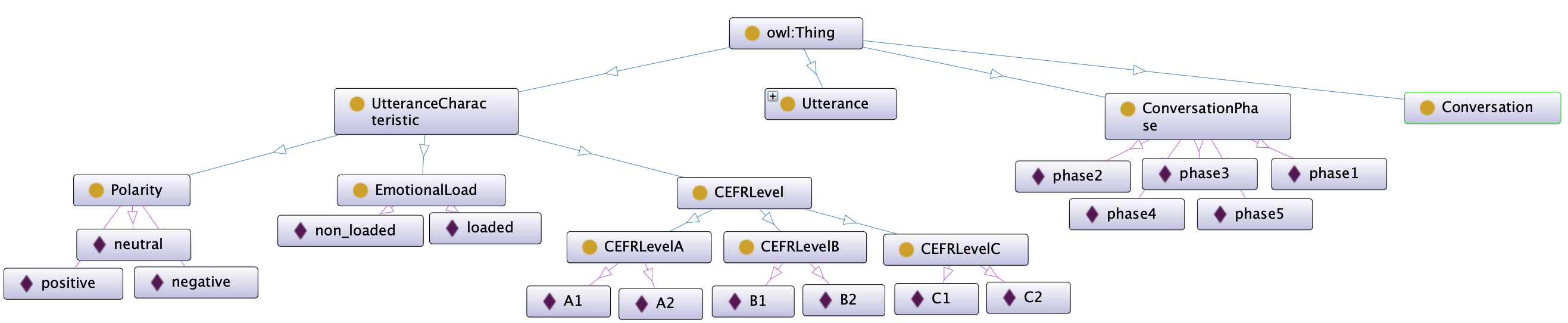}
    \caption{\centering OntoGraf~\cite{falconer2010ontograf} visualization of the general conversation ontology and its links to the specific concepts used in our conversation strategy.}
    \label{fig:onto-control-aspects}
\end{figure*}

This section describes both the methodology and the experimental setup of our contribution. It first provides an overview of the ontological framework devised for the definition of conversation strategies, and then details how the use-case is defined and implemented in this framework. Finally, we describe the evaluation process relying on this use-case.

\subsection{Ontology Design and Use-Case Description}

In order to enable re-usability, the ontology is built following different levels of abstraction. As the most general layer, we focus on the foundational aspects of the conversation. In this way, we conceptualize the notion of conversation and its evolution. Here, we consider only dyadic user-agent textual conversations. Such a conversation, which is an instance of the \texttt{Conversation} concept, contains utterances, so we create an \texttt{Utterance} concept, divided into sub-concepts depending on whether the utterance originates from the user or the agent. To represent the evolving nature of conversation, we define the notion of conversation phase through the \texttt{ConversationPhase} concept, with a property linking an utterance with the conversation phase to which it belongs. In addition, we define the \texttt{UtteranceCharacteristic} concept, which allows us to link an utterance to characteristics useful for the definition of conversation strategies, using object or data properties. This type of utterance characterization, which is then specific to the strategy under consideration, will also make it possible to characterize different phases of the conversation. We propose a first general layer of conceptualization of the conversation (the first depth level in Figure~\ref{fig:onto-control-aspects}) which offers a formal basis for defining use-case-specific conversation strategies. Finally, to guide the LLM with respect to the content to be generated, each phase is associated with a system prompt passed to the LLM and that is specific to the phase.

We illustrate the conversation strategy design through the use-case of job interview practice. The idea is to allow the user to interact with an agent that plays the role of a recruiter, asking for multiple details and questions on various aspects of the user-candidate profile. More precisely, in order for the user to experience several conversation scenarios, we propose a 5-phase conversation, delimited in terms of the number of user utterances per phase:
\begin{itemize}
    \item Phase 1: small talk at the beginning of the interview;
    \item Phase 2: the agent asks the user to present themselves along with their background and how it relates to the job position;
    \item Phase 3: technical discussion on a topic chosen by the agent, that is connected with the user's area of expertise and the job position;
    \item Phase 4: they debate a controversial topic chosen by the agent;
    \item Phase 5: the agent sums up and concludes the interview.
\end{itemize}

At this point, we need to define more precisely what should happen at each conversation turn: how the agent should respond according to the context, and which characteristics of its responses should evolve according to which characteristics of the user utterances, and the current phase. To this end, we define descriptors that describe the characteristics to be controlled, at utterance level. In our case, the descriptors used are the utterance's language level CEFR (using the 6-level scale provided by the Common European Framework of Reference for Languages\footnote{\url{https://www.coe.int/en/web/common-european-framework-reference-languages/table-1-cefr-3.3-common-reference-levels-global-scale}}), its emotional load $L$ (0=uncharged, 1=emotionally charged) and its emotional polarity $P$ (0=negative, 1=neutral, 2=positive). We illustrate the enriched version of the ontology for this specific use-case in Figure~\ref{fig:onto-control-aspects}.

To obtain the descriptor values from the user utterances, we leverage models that we previously trained on adequate corpora for each descriptor. The load and the polarity are predicted using Transformer encoder language models~\cite{vaswani2017attention} based on the BERT architecture~\cite{devlin2019bertpretrainingdeepbidirectional}. Regarding the CEFR level, we extracted 44 linguistic features and trained a decision tree on CEFR-level classification in order to obtain rules of belonging to each class depending on the most important linguistic features. We then converted such definitions into description logics to incorporate them in the ontology. Hence, the CEFR level of an utterance is determined directly \textsl{via} ontological reasoning.
Depending on the values taken by these descriptors, we define different sub-classes of \texttt{Utterance} useful for implementing the strategy. For example, an instance of \texttt{MaxB\_UserUtterance} is a user utterance for which its CEFR level, that of the previous one or that of the one before is B1 or B2, and none of the three is C1 or C2 (i.e. over the last three turns, the users expressed themselves to at most a B level). 

These subclasses are used to further characterize the five conversation phases through general class axioms. For instance, the axiom below states that a \texttt{MaxB\_UserUtterance} encountered in phase 2 leads to the next agent utterance being at A2 level.

\begin{center}
    \texttt{MaxB\_UserUtterance \textcolor{cyan}{\bf and} PhaseTwoUtterance \textcolor{blue}{\bf SubClassOf} nextCharacteristic \textcolor{violet}{\bf value} A2}
\end{center}

Descriptors not only provide a formal representation of utterance-level characteristics but especially serve as control signals in our hybrid system. The ontology plays a central role as it defines the semantics of the control space, while the LLM needs to be fine-tuned to generate responses aligned with descriptor values. In this way, it will be possible to establish ontology-driven conversational control.

\subsection{Evaluation Strategy}

Given our system, two evaluation phases are involved throughout the process. First, we must verify that the control signals -- \textit{i.e.} the descriptor values -- can indeed be used to guide generation. Since these descriptors are defined within the ontology, their representations must be learned by the LLM, which motivates the use of fine-tuning. At this stage, we only need to generate content under the constraint of one or more randomly selected descriptor values. It is not necessary for these values to be consistent with the conversation strategy, since we are not building up a dialogue yet. We therefore fine-tune the LLM in a zero-shot setting for controlled generation. Moreover, in order to prompt relevant text generation, we include a randomly selected open-ended question in the input. For this purpose, we designed a list of questions likely to elicit responses whose properties span the full range of descriptor values, thus enabling a relevant learning of the different concept combinations. Further details about our fine-tuning setup and the results of this first evaluation phase are given in Section~\ref{sec:ft}. 

Afterwards, starting from this language model trained to generate content according to descriptor values, we need to evaluate compliance with the conversation strategy as a whole, and with that judge the overall suitability of our conversational agent for real users. To this end, human evaluators each engaged in several conversations with the agent. We assessed the quality of the interactions both in terms of objective compliance with the strategy and the evaluators’ subjective perceptions of coherence and appropriateness regarding the context. Thus, we follow a perspectivist approach for the evaluation~\cite{frenda2024perspectivist,nlperspectives-2024-perspectivist}. Further details about our in-conversation experiments and evaluation are given in Section~\ref{sec:conv}.

\section{Fine-Tuning Experiments}
\label{sec:ft}

\subsection{Experimental Setup}

We use fine-tuning to control the LLM generation over the conversation aspects defined in the ontology: the proficiency level, the emotional load, and the emotion polarity. This is a crucial step in order for the model to be able to comply with the strategy when in an actual conversation. Hence, we consider the framework of constraint generation, where we need to first define how the constraint is given to the model, and then rely on an appropriate learning procedure. The constraint is defined in a control code, as introduced by Keskar \textit{et al.}~\cite{keskar:2019}, that simply consists of adding a prefix upwind the generated content that indicates, for each aspect to control, the ontology concept name and its value. For instance, \texttt{[ProficiencyLevel: A2, Polarity: positive]} and \texttt{[Load: 0]} are valid prefixes.

Regarding learning, we avoid using annotated data, as the targeted aspects are subjective and context-dependent, making human annotation biased. Constructing a dataset that adequately spans our multi-label annotation strategy would require extensive, fine-grained supervision prone to noise and selection bias. Therefore, we instead adopt a data-free learning approach based on reinforcement learning, following recent advances in language model training for reasoning~\cite{deepseekai2025deepseekr1incentivizingreasoningcapability}. We employ Group Relative Policy Optimization -- GRPO~\cite{shao2024deepseekmathpushinglimitsmathematical} --, directly inspired by Proximal Policy Optimization -- PPO~\cite{PPO} --, which presents better stability and convergence since the reward is aggregated over a group of samples. In our case, this method proves efficient in improving compliance with multiple aspects of the content, while it has proven challenging using standard supervised fine-tuning, often resulting in catastrophic forgetting~\cite{dong-etal-2024-abilities}. We use adapter layers, which are small additional layers inserted in the model~\cite{houlsby}, and more specifically LoRA (Low-Rank Adaptation) adapters~\cite{lora-adapters}, which inject trainable low-rank matrices into attention layers. This allows parameter-efficient fine-tuning, as only a small fraction of weights are updated. We make our trained LoRA adapters available to the community for the sake of reproducibility.

Our fine-tuning procedure consists of providing prompts with a randomly sampled question (that plays the role of the user utterance), and giving randomly sampled instructions for descriptor values in a prefix. The GRPO reward function evaluates the compliance of the generated content with the given instructions.

\subsection{Results}

To validate our approach, we compare two state-of-the-art language models in their instruction-tuned versions, namely Llama3.1-8B-Instruct~\cite{dubey2024llama3herdmodels} and Qwen2.5-7B-Instruct~\cite{yang2024qwen2technicalreport}. Instruction tuning enhances the model’s responsiveness to prompts, making it possible to design system prompts that clearly describe the task and incorporate both the strategy guidelines and the ontology descriptor values. We focus on relatively lightweight models of comparable sizes (7B and 8B parameters). This is primarily due to computational constraints, but it is also motivated by the will to demonstrate the effectiveness of ontology-based control even in smaller models. If such models can successfully adapt their behavior along ontological specifications, it suggests that our approach is viable in low-resource settings, without requiring much larger architectures to observe a significant change in behavior. Controlling models is also more important for smaller ones, as we can derive emerging abilities from them with fewer parameters to adjust compared to larger models (i.e., more than 70 billion parameters)~\cite{lu-etal-2024-emergent}.

Our results are presented in Table~\ref{tab:ft_res}. Although several aspects may be controlled simultaneously during one fine-tuning iteration, we present the results by descriptor in order to compare the language model's integration of the different concepts. In general, our metric values suggest that fine-tuning improves compliance with the control instructions, while the task itself is non-trivial as it often requires the simultaneous handling of multiple independent descriptors. We also observe that the generated responses generally align with the targeted descriptor values, confirming the feasibility of our control framework. Furthermore, in addition to usual classification metrics, we provide the Mean Average Error -- MAE~\cite{mae:2005} -- and the Matthews Correlation Coefficient -- MCC~\cite{MATTHEWS1975442} --. MAE is necessary for a fair assessment of proficiency level control, due to the ordinal nature of CEFR levels. Otherwise, MCC accounts for prediction relevancy, as it consists of a class-wise Pearson correlation between actual and predicted samples. Regarding these additional metrics, we mostly observe improvements as well. Nevertheless, challenges remain, especially regarding the control of the emotional load. Indeed, most of the generated content is emotionally loaded, likely because of the emphasis placed by the developers of pre-trained models on designing models that do not sound artificial and provide pleasing discussions. Also, instruction-tuned models often display a bias toward overly positive or neutral tones due to their objectives during the instruction tuning step.

\begin{table*}[]
\setlength{\tabcolsep}{4pt} 
\centering
\begin{tabular}{@{}c ccc|ccc|ccc@{}}
\toprule
\multirow{2}{*}{\bf Model} & \multicolumn{3}{c|}{\bf Proficiency level} & \multicolumn{3}{c|}{\bf Emotional load} & \multicolumn{3}{c}{\bf Emotion polarity} \\ \cmidrule(l){2-10}
                       & Acc & wF1 & MAE & Acc & wF1 & MCC & Acc & wF1 & MCC \\ \midrule
\multicolumn{10}{c}{\sl Pre-trained baselines} \\ \midrule
Llama3.1-8B-Instruct   & 0.18 & 0.15 & 1.92 & 0.50 & 0.36 & 0.005 & 0.42 & 0.39 & 0.14 \\
Qwen2.5-7B-Instruct    & 0.18 & 0.16 & 1.77 & 0.49 & \textbf{0.48} & -0.02 & 0.46 & 0.46 & 0.20 \\ \midrule
\multicolumn{10}{c}{\sl Ours: GRPO fine-tuning on ontology concepts} \\ \midrule
Llama3.1-8B-Instruct   & \textbf{0.29} & \textbf{0.26} & \textbf{1.31} & \textbf{0.51} & 0.38 & 0.009 & \textbf{0.62} & \textbf{0.58} & 0.15 \\
Qwen2.5-7B-Instruct    & 0.21 & 0.20 & 1.64 & 0.46 & 0.41 & -0.10 & 0.57 & 0.56 & \textbf{0.40} \\ \bottomrule
\end{tabular}
\caption{\centering Evaluation of our fine-tuning procedure for controlled generation according to the ontology descriptor values. We compare the control abilities of each LLM with its associated pre-trained baseline. wF1 refers to the weighted F1 score.}
\label{tab:ft_res}
\end{table*}

\section{In-Conversation Experiments}
\label{sec:conv}

\subsection{Experimental Setup}

After evaluating the impact of ontology-based control learned through fine-tuning on language models, we propose a methodology for assessing the conversational agent itself. To this end, we conduct full conversations to test the agent's adherence to the complete set of rules defined by the conversation strategy. Each conversation consists of 12 dialogue turns (i.e., one user utterance followed by one agent utterance), structured according to the five phases described in Section~\ref{sec:approach}. Three human evaluators, with an academic background in computer science, interacted with the agent by role-playing as interviewees, adopting different levels of experience and varying areas of expertise. This setup was designed to validate the applicability independently from the specific topic or domain of the conversation. We evaluate the quality of the agent based on two main criteria. The first is strict compliance with the conversation strategy as evaluated through the classifiers of descriptors: for each agent utterance, we compare the requested characteristics derived from the ontology with those detected in the response by those classifiers -- the two models for polarity and emotional load, and the ontology itself for language level. 
However, those descriptors are subjective, and classifiers might focus on a different part of the utterance than a human would in the context of the scenario, especially when the utterance starts with an emotionally charged polite expression, for instance. Thus, the second criterion is based on human judgment: each evaluator annotated the agent's utterances with the characteristics they perceived, without access to the expected or detected values, nor to the annotations from the other evaluators. 

Since the CEFR level of each utterance is formally inferred through ontological reasoning, its annotation does not rely on subjective interpretation. Consequently, the human evaluators focused solely on the perceptual descriptors—emotional load and polarity—whose assessment requires subjective judgment and cannot be objectively derived from linguistic features alone.
All evaluators independently annotated the full set of conversations.

\subsection{Results}

First, we evaluated whether the agent followed the five-phase structure of the job interview scenario. Here, instruction-tuned LLMs demonstrated a strong ability to respond appropriately to phase transitions triggered by system prompts. Despite limitations in fine-grained stylistic control, the agent reliably maintained the structure and content required for each interview phase, reacting coherently to the evolving context. In particular, transitions -- such as moving from small talk to technical discussion -- were handled smoothly, as illustrated in Figure~\ref{fig:conv_ex}. This figure illustrates examples of all possible phase transitions in our conversational setup, drawn from the collected conversation corpus. The examples presented in the paper were selected to be representative of the typical behavior observed across most conversations in the corpus.

\begin{figure*}[!ht]
    \centering
    \begin{tcolorbox}[colback=gray!5, colframe=white, boxrule=0mm, sharp corners, left=1.5mm, right=1.5mm]
        \begin{tabbing}
            \textcolor{blue}{\bf Exit small talk}\\[0.3em]
            \hspace{1.5cm} \= \hspace{10cm} \= \kill 
            \textbf{User:} \> \parbox[t]{10cm}{\textit{Sure. [...] I used to be a data science intern and now I apply to my first full time position as a data scientist.}} \\[0.3em]
            \textbf{Agent:} \> \parbox[t]{10cm}{That's great! [...] Can you tell me more about your experience as a data science intern? What kind of projects did you work on during that time, and what skills did you develop?\enspace (\textcolor{teal}{\bf $L=0$, $P=2$} / \textcolor{orange}{B1, $L=1$, $P=2$})} \\[0.8em]
            \textcolor{blue}{\bf Switch from phase 2 (background) to phase 3 (technical point)}\\[0.3em]
            \textbf{User:} \> \parbox[t]{10cm}{\textit{I have real experience and I know how to adapt quickly to many different contexts and technical stacks. Also, if your company has a home-made framework, I can learn it quite fast.}} \\[0.3em]
            \textbf{Agent:}  \> \parbox[t]{10cm}{Let's dive into the technical aspects [...]. [...] How do you handle debugging and optimizing the performance of a cross-platform desktop application built with Electron [...] ?\enspace (\textcolor{teal}{\bf $L=0$} / \textcolor{orange}{C1, $L=1$, $P=1$})} \\[0.8em]
            \textcolor{blue}{\bf Switch from phase 3 (technical point) to phase 4 (controversial topic)}\\[0.3em]
            \textbf{User:} \> \parbox[t]{10cm}{\textit{The data came directly from the personal tool of each worker. [...] It turns out that, before drift, this was enough to get low errors.}} \\[0.3em]
            \textbf{Agent:} \> \parbox[t]{10cm}{That's a good approach [...]. [...] However, some might argue that relying on human performance data [...] might not be the most reliable method. How do you respond to that criticism?\enspace (\textcolor{teal}{$L=1$, $P=2$} / \textcolor{orange}{B1, $L=1$, $P=2$})} \\[0.8em]
            \textcolor{blue}{\bf Reach the conclusion of the interview}\\[0.3em]
            \textbf{User:} \> \parbox[t]{10cm}{\textit{Yes, I am looking for ways to improve and optimise. [...]}} \\[0.3em]
            \textbf{Agent:} \> \parbox[t]{10cm}{It seems like we've had a good discussion about your experience and skills. I'd like to summarize what we've covered so far. [...] You've also demonstrated your ability to approach complex problems and think critically about solutions.\enspace (\textcolor{teal}{B1, $L=0$, $P=1$} / \textcolor{orange}{C1, $L=1$, $P=2$})} 
        \end{tabbing}
    \end{tcolorbox}
    \caption{\centering Illustrations of the conversational agent abilities to switch between the phases, leave the small talk at the beginning and conclude in the end. At the end of each agent utterance, we include the \textcolor{teal}{expected} / \textcolor{orange}{detected} descriptor values.}
    \label{fig:conv_ex}
\end{figure*}

While the examples show a good compliance to the phases, they also highlight that the model is not adaptable enough regarding style constraints (emotional load, polarity and language level). It is partly understandable from the previous part of our process (the fine-tuning) since we only manage to enable a light control on the load, for instance. 
To further assess the fine-tuned model’s compliance with the ontology-based constraints, the three human annotators conducted conversations with both the pre-trained and fine-tuned versions of the model. These experiments were performed using Llama 3.1, which, as previously noted, achieved the best fine-tuning performance.
The first row of Table~\ref{tab:conv_res} presents the accuracy score between the expected and detected descriptor values across all the agent utterances of the whole corpus of conversations. The score values confirm this lack of adaptability of the language model, even after fine-tuning. Nevertheless, from a qualitative point of view, the agent's utterances seem to generally align with what a human user might reasonably expect. 
One possible explanation for this discrepancy between expected and detected scores lies in the overly frequent use of polite or encouraging expressions. Most generated utterances begin with such sentences, which tends to bias automatic predictions of load and polarity toward a loaded positive characterization of the whole utterance. As a result, the descriptor values perceived by human evaluators are often closer to the expected values than to those detected by automatic analysis, as shown in Table~\ref{tab:conv_res}.

Regarding the comparison between the pre-trained and fine-tuned models, our findings reinforce the idea that learning to control emotional load remains particularly challenging. Indeed, for most annotators, the pre-trained model exhibits higher compliance with the load constraint. In contrast, polarity—which had already shown responsiveness to the fine-tuning procedure—confirms this trend, as most annotators report that the fine-tuned model aligns more consistently with the specified polarity constraint.

In summary, the conversational agent reliably follows the scenario structure but shows limited control over stylistic aspects. The discrepancy between expected and detected values highlights the difficulty of fine-grained conditioning, though human judgments suggest a closer alignment with intended behavior than accuracy scores between expected and detected descriptor values.

\begin{table*}[ht!]
\centering
\setlength{\tabcolsep}{4pt}
\begin{tabular}{@{\hskip 2pt}l@{\hskip 12pt}cc@{\hskip 12pt}cc@{\hskip 2pt}}
\toprule
 & \multicolumn{2}{c}{\textbf{Pre-trained}} & \multicolumn{2}{c}{\textbf{Fine-tuned}} \\ 
\cmidrule(lr){2-3} \cmidrule(lr){4-5}
\textbf{Accuracy score} & \textbf{Load} & \textbf{Polarity} & \textbf{Load} & \textbf{Polarity} \\ 
\midrule
Expected vs detected         & \textbf{0.24} & 0.34 & 0.18 & \textbf{0.40} \\
Expected vs A\#1             & \textbf{0.74} & 0.27 & 0.67 & \textbf{0.30} \\
Expected vs A\#2             & \textbf{0.85} & 0.32 & 0.64 & \textbf{0.45} \\
Expected vs A\#3             & 0.63 & \textbf{0.35} & \textbf{0.77} & 0.33 \\
Inter-annotator agreement    & 0.81$\pm$0.05 & 0.74$\pm$0.02 & 0.78 $\pm$ 0.03 & 0.66 $\pm$ 0.02 \\
\bottomrule
\end{tabular}
\caption{\centering Comparison of accuracy scores between quantitative and human evaluation. We report individually the scores of our three annotators (A\#1, A\#2, A\#3), along with a pairwise inter-annotator agreement score for each descriptor. We compare Llama3.1-8B-Instruct pre-trained and fine-tuned. Best scores are in bold.
}
\label{tab:conv_res}
\end{table*}

\section{Limitations}

\paragraph{Overly Positive Generation Bias.}
The limitations of current pre-trained LLMs pose challenges to our approach. Indeed, they tend to generate excessively positive or flattering responses—also referred to as sycophancy~\cite{sicilia-etal-2025-accounting}—which constrains the stylistic and affective variations we aim to induce through descriptors such as emotional load and polarity. This behavior largely results from post-training alignment steps designed to prevent harmful or overly negative outputs, but it also makes the models reluctant to adopt more neutral or critical tones. While our system successfully guides the overall structure and content of the conversation, these stylistic biases remain difficult to overcome. Future work should therefore investigate more targeted fine-tuning or stronger alignment with ontological control signals to attenuate this overly positive tendency.

\paragraph{Ontology Requirement.}
Another limitation concerns the need for a dedicated ontology to represent the relevant descriptors and conversation strategies. Although this is the only component that must be adapted, ontology design is not always trivial and may involve a non-negligible knowledge modeling effort, especially for domain-specific applications. This requirement slightly limits the plug-and-play nature of our framework compared to other approaches that do not rely on structured representations~\cite{goswamy2020adapting}. Nonetheless, this reliance on an ontology also provides an interpretable and systematic foundation for integrating domain knowledge, which we consider a key advantage for controlled conversational generation.

\paragraph{Human Evaluation.}
When working on conversational agents, it is crucial to rely on a diverse panel of external human evaluators to ensure a fair and unbiased assessment of the proposed system. In our case, it is essential to determine whether users tend to perceive the response generated by the fine-tuned model or that of the pre-trained model as better aligned with the ontology-defined control signal. This feedback would allow us to verify that the intended conversational control is indeed perceived by users and therefore effectively implemented. To this end, we have developed a demonstrator designed to facilitate large-scale human evaluation involving a broad range of annotators, but this evaluation is still to be conducted.

\section{Conclusion and Future Work}

We propose an ontology-driven framework for controlling conversational language models, with a specific focus on open-domain dialogues and exploring the use-case of job interview practice. Our results demonstrate that small language models (7B and 8B), when guided by a formally structured ontology in the prompt, can reliably follow complex multi-phase conversation strategies and adapt their responses according to multiple conversational descriptors. Crucially, our framework does not require modifying the base model architecture: it operates by combining reasoning over ontological descriptors with prompt-level control codes. This hybrid approach leads to a lightweight conversational agent that follows a high-level dialogue strategy while remaining flexible across use-cases. Redefining the ontology would allow the same framework to be reused as such for completely different scenarios, hence requiring minimal engineering. Overall, this work highlights the viability of ontological control as a foundation for developing transparent, strategy-aware conversational agents. As LLMs become more adaptable and controllable, our framework allows for predictable conversation flows while being able to scale, therefore offering an interpretable and modular approach to personalized dialogue generation.

In that direction, future work would explore other use-cases that may necessitate more complex conversation strategies. This requires designing new ontologies as well as representing more descriptors in order to enable control on more aspects of the conversation. This increasing complexity of conversational strategies may require scaling up to larger LLMs with greater parameter capacity, allowing them to assimilate and represent more intricate conceptual relationships. To achieve this, reinforcement learning procedures such as GSPO -- Group Sequence Policy Optimization, a recent improvement of GRPO --~\cite{zheng2025groupsequencepolicyoptimization} will be necessary to maintain the maximum efficiency of our approach.

\paragraph*{Supplemental Material Statement:} In order to transparently share details about our framework and include all our results, we share the following as supplementary material:
\begin{itemize}
    \item Our conversation ontology that also contains the strategy for job interview practice;
    \item Source code to train CEFR levels, load and polarity predictors, in addition to the predictors themselves when they are small enough;
    \item Source code to fine-tune LLMs using GRPO on our generation control task along with the associated trained adapter layers; 
    \item Source code to interact with the conversational agent, the 30 conversations done by the human evaluators along with evaluation results.
\end{itemize}

%
%
%
\bibliographystyle{splncs04}
\bibliography{references}
\end{document}